\documentclass{article}
\usepackage[a4paper, total={6in, 8in}]{geometry}
\usepackage{hyperref}
\usepackage{url}            
\usepackage{graphicx}
\usepackage{tabularx,ragged2e}
\usepackage{algorithm,algpseudocode}
\usepackage{natbib}
\usepackage[symbol]{footmisc}

\usepackage{amsfonts,mathtools}
\usepackage{multirow}
\usepackage{mathtools}
\newcommand{\E}{\mathbb{E}}
\newcommand{\sm}{\text{Softmax}}

\begin{document}
\begin{center}
	\LARGE Soft Actor-Critic With Integer Actions 
\end{center}
\vspace{0.3cm}

\centerline{\large Ting-Han Fan\textsuperscript{\rm 1,2} and Yubo Wang \textsuperscript{\rm 2}} 
\vspace{0.3cm}
\centerline{\large \textsuperscript{\rm 1} Princeton University}
\centerline{\large \textsuperscript{\rm 2} Siemens Technology}
\vspace{0.3cm}
\centerline{\large \texttt{tinghanf@princeton.edu, yubo.wang@siemens.com}}
\vspace{0.3cm}

\begin{abstract}
Reinforcement learning is well-studied under discrete actions. Integer actions setting is popular in the industry yet still challenging due to its high dimensionality. To this end, we study reinforcement learning under integer actions by incorporating the Soft Actor-Critic (SAC) algorithm with an integer reparameterization. Our key observation for integer actions is that their discrete structure can be simplified using their comparability property. Hence, the proposed integer reparameterization does not need one-hot encoding and is of low dimensionality. Experiments show that the proposed SAC under integer actions is as good as the continuous action version on robot control tasks and outperforms Proximal Policy Optimization on power distribution systems control tasks.
\end{abstract}

\section{Introduction}
Recent breakthroughs of Reinforcement learning (RL) have shown promising results from Atari games \citep{mnih2013atari}, MuJoCo continuous control \citep{todorov2012mujoco}, robotics \citep{levine2016robotics}, to autonomous driving \citep{highway-env}. However, in industrial scenarios such as power systems \citep{marot2021power-env}, resource allocation \citep{zoltners1980integer}, and scheduling \citep{wang2005scheduling}, the decision variables are often represented by integer quantities. Because the integer variables are not easily differentiable, they raise an issue to the DDPG-style policy gradient \citep{silver2014ddpg,haarnoja2018sac,fujimoto2018td3}, a state-of-the-art RL method. While the REINFORCE-based methods \citep{glynn1990reinforce,schulman2015trpo,schulman2017ppo} are still alternatives to the integer problem, they tend to have high variance \citep{Schulman2015advantage,wu2018variance} and have been shown to be outperformed by the DDPG-style methods \citep{haarnoja2018sac,fujimoto2018td3}. Therefore, it is of practical interest to study methods to parameterize integer random variables and apply them to DDPG-style RL methods.

Gumbel-Softmax is a state-of-the-art reparameterization method for discrete random variables \citep{Jang2017gumbel,Chris2017gumbel}. Combining with the Straight-Through estimators \citep{bengio2013st,Chung2017st}, the Straight-Through Gumbel-Softmax parameterization can represent a discrete random variable in the forward pass (of a neural network) and maintain differentiability in the backward pass. Since integer variables admit a discrete structure, it is appealing to apply the Gumbel trick for our integer problems.

Our key insight is that integer and discrete variables are fundamentally different: integer variables are comparable while the discrete variables aren't in general. For example, the elements in the integer set $\{1,2,3\}$ are comparable because $1<2<3$, while the discrete set $\{\text{dog},\text{cat},\text{frog}\}$ can't assign such a relation. Recall that the one-hot encoding is required for representing a discrete sample due to its incomparability. An integer set has comparability. Ideally, we only need an integer, not a one-hot vector, to represent an integer sample. In light of this fact, we propose a reparameterization for an integer random variable: use a linear mapping from a one-hot vector of Straight-Through Gumbel-Softmax to an integer. The mapped integer is random and differentiable (w.r.t. parameters of a neural network) because the one-hot vector is. Therefore, this is a valid reparameterization for integer random variables.

Another advantage of the integer-mapping trick is that the integer space can be mapped back to the ambient space, making the effective dimension irrelevant to the fineness of discretization. For example, suppose the ambient action space is an interval $\mathcal{A}=[-1,1]$, and the interval is discretized into $N\geq 2$ points. Then the discretized action space $\mathcal{A}_d=\{0,..,N-1\}$ is mapped back to the ambient action space using $\mathcal{T}(\mathcal{A}_d)=\mathcal{A}_d\frac{2}{N-1}-1$. Since $\mathcal{T}(\mathcal{A}_d)\subset [-1,1]$, the effective dimension of $\mathcal{T}(\mathcal{A}_d)$ remains to be one and irrelevant to the fineness of discretization $N$.

With (1) the differentiability and (2) the advantage on the effective dimension of action space, the proposed integer reparameterization is particularly useful for the DDPG-style method such as the Soft-Actor Critic (SAC) \citep{haarnoja2018sac}. Thereby, we propose a variant of SAC under integer actions by incorporating our integer reparameterization into the actor network (policy) with the critic network (Q function) staying the same. This is possible because (1) allows us to differentiate through actions and (2) makes the input/output dimension of the actor and critic networks the same as that of the continuous action version.

We validate the proposed SAC with integer actions on PyBullet robot control \citep{coumans2021bullet} and PowerGym power systems control \citep{fan2021powergym}. The results show the proposed SAC under integer actions achieve similar performance as the continuous action counterpart in PyBullet and can outperform the PPO \citep{schulman2017ppo} algorithm in PowerGym. Hence, this is a certificate of the effectiveness of our SAC with integer actions.

\section{Reinforcement Learning}
We consider an infinite-horizon Markov Decision Process (MDP) $\langle\mathcal{S},~\mathcal{A},~ T,~r,~\gamma\rangle$. $\mathcal{S},~\mathcal{A}$ are state and action spaces, $r(s,a,s')$ is the reward function, $\gamma\in(0,1)$ is the discount factor, and $T(s'|s,a)$ is the state transition density. Let the initial state distribution be $\rho_0$. The objective of RL is to find a policy $\pi$ that maximizes the $\gamma$-discounted cumulative reward:
\begin{equation}
\max_\pi J(\pi) = \max_\pi \E\left[\sum_{i=0}^\infty  \gamma^i r(s_i,a_i)\Big\lvert s_0\sim\rho_0,~a_i\sim\pi(\cdot|s_i),~s_{i+1}\sim T(\cdot|s_i,a_i)\right].
\label{eq:rl_obj}
\end{equation}
Suppose the policy is parameterized by $\theta$ as $\pi_\theta$. A typical way to optimize Eq.~\eqref{eq:rl_obj} by tuning over $\theta$ is to use the policy gradient, which depends on the Q function under $\pi_\theta$:
\begin{equation}
Q^{\pi_\theta}(s,a)=\mathbb{E}\left[\sum_{i=0}^\infty\gamma^i r(s_i,a_i)\Big\lvert s_0=s,~a_0=a,~a_i\sim\pi_\theta(\cdot|s_i),~s_{i+1}\sim T(\cdot|s_i,a_i)\right].
\label{eq:qfunc}
\end{equation}
With $Q^{\pi_\theta}$ defined above, the policy gradient theorem~\citep{sutton2000policy} states that
\begin{equation}
    \nabla_\theta J(\pi_\theta)=\E_{s\sim\rho^\pi,~a\sim\pi_\theta}[\big(\nabla_\theta \ln \pi_\theta(a|s)\big)Q^{\pi_\theta}(s,a)],
\label{eq:pg}
\end{equation}
where $\rho^\pi$ is the unnormalized discounted state distribution. Since the expectation of Eq.~\eqref{eq:pg} cannot be exactly computed, the gradient estimation becomes an important problem. We will review this topic specifically on discrete random variables in the next section.

\section{Gradient Estimation for Discrete Random Variables}
Let $D$ be a random variable from a discrete distribution with $N\geq 2$ elements. Let $S=\{s_1,...,s_N\}$ be the support of this discrete distribution. Without loss of generality, suppose the discrete distribution $p_\theta$ is parameterized by a neural network with parameter $\theta$. We can write
$$D\sim p_\theta,\quad\sum_{d\in S}p_\theta(d)=1.$$
Given an objective function $f:S\rightarrow\mathbb{R}$, we minimize
$$\min_\theta \E[f(D)],$$
where the expectation is taken over the random sample $D$. This can be done by computing some estimates of the gradient $\nabla_\theta \E[f]$, which we will review in this section.

\subsection{REINFORCE Estimator}
The REINFORCE estimator \citep{glynn1990reinforce,williams1992reinforce} utilizes a fact from the chain rule:
$$\nabla_\theta p_\theta(d)=p_\theta(d)\nabla_\theta \ln p_\theta(d)$$
Therefore, $\nabla_\theta \E[f]$ is reformulated as
$$\nabla_\theta \E[f]=\sum_{d}\nabla_\theta p_\theta(d)f(d)=\sum_{d}p_\theta(d)\nabla_\theta \ln p_\theta(d)f(d)=\E[\nabla_\theta\ln p_\theta(D)f(D)],$$
and the REINFORCE estimator simply takes Eq.~\eqref{eq:grad-reinforce} as an unbiased estimate of $\nabla_\theta \E[f]$.
\begin{equation}
    \nabla_\theta\ln p_\theta(D)f(D)
\label{eq:grad-reinforce}
\end{equation}
Since the policy gradient Eq.~\eqref{eq:pg} and the REINFORCE estimator Eq.~\eqref{eq:grad-reinforce} share a similar form, it is tempting to develop a policy gradient algorithm upon the REINFORCE estimator. Examples of such algorithms include TRPO \citep{schulman2015trpo} and PPO \citep{schulman2017ppo}. With the REINFORCE estimator, TRPO and PPO have high variances and however do not require differentiable random actions. On the contrary, without the REINFORCE estimator, the DDPG-style methods have lower variances but require differentiable random actions. Hence, to reduce the variance, we want to build models for differentiable random actions and use DDPG-style methods such as the SAC to tackle our integer-constrained problem.

\subsection{Gumbel-Softmax}
Gumbel-Softmax \citep{Jang2017gumbel,Chris2017gumbel,paulus2021raoblackwellizing} provides an alternative solution of discrete random variable modeling. By introducing the softmax relaxation, it is considered to be of lower variances during the gradient computation. 

Concretely, define the softmax function $\sm:\mathbb{R}^n\rightarrow\mathbb{R}^n$ as $[\sm(x)]_i=e^{x_i}/\sum_{j=1}^ne^{x_j}$ and suppose $\text{logit}_\theta\in \mathbb{R}^n$, the unnormalized log probability vector, is parameterized by $\theta$. The Gumbel-Max trick \citep{chris2014gumbel} states that
\begin{equation}
D\sim \sm(\text{logit}_\theta)\overset{d}{=}\arg\max (\text{logit}_\theta+G),
\label{eq:gumbel-max}
\end{equation}
where G is an i.i.d. $n$-dimensional Gumbel(0,1) vector. $D\sim \sm(\text{logit}_\theta)$ means that $D$ is sampled from the discrete probability distribution formed by $\sm(\text{logit}_\theta)$. The notation $\overset{d}{=}$ means the left and the right have the same distribution.

Since Eq.~\eqref{eq:gumbel-max} is not differentiable (both the sampling and the argmax), the Gumbel-Softmax approximates the $\arg\max$ of Eq.~\eqref{eq:gumbel-max} using a Softmax function, which is differentiable.
\begin{equation}
    D_{\text{GS}}=\sm (\text{logit}_\theta+G)
\label{eq:gs}
\end{equation}
Despite bringing back the differentiability, Eq.~\eqref{eq:gs} inevitably introduces a bias since $D_{\text{GS}}$ is not an one-hot vector. This can be solved by the Straight-Through estimator \citep{hinton2012neural}[lecture 15b]. Let $D_{\text{hot}}=\text{OneHot}(D)$ be the one hot version of $D$ from the $\arg\max$ of Eq.~\eqref{eq:gumbel-max}. The Straight-Through Gumbel-Softmax \citep{Jang2017gumbel} introduces a constant shift of $D_{\text{GS}}$:
\begin{equation}
    D_{\text{STGS}}=D_{\text{hot}}-[D_{\text{GS}}]_{\text{const}}+D_{\text{GS}},
\label{eq:stgs}
\end{equation}
where $[D_{\text{GS}}]_{\text{const}}$ treats $D_{\text{GS}}$ as a constant by detaching the dependency on $\theta$ during the backward propagation of a neural network. Both $D_{\text{hot}}$ and $[D_{\text{GS}}]_{\text{const}}$ are constants for derivatives w.r.t. $\theta$ are zeros. Therefore, $D_{\text{STGS}}$ is one-hot in the forward propagation (of a neural network) and differentiable in the backward propagation. Redefine the objective function as a differentiable $f:\mathbb{R}^n\rightarrow\mathbb{R}$. The gradient estimate under $D_{\text{STGS}}$ becomes
\begin{equation}
   \nabla_{\theta} f(D_{\text{STGS}})=\bigg[\frac{\partial f(D_{\text{STGS}})}{\partial \theta_i}\bigg]_{i},
\label{eq:grad-stgs}
\end{equation}
where $\theta_i$ is the ith element of the model parameter $\theta$. The Straight-Through Gumbel-Softmax estimator is believed to be of lower variance, prevents error in forward propagation, and maintains sparsity \citep{bengio2013st,Chung2017st,Jang2017gumbel}. Hence, Eq.~\eqref{eq:grad-stgs} has become a popular alternative of Eq.~\eqref{eq:grad-reinforce}, and we would like to incorporate it into the SAC algorithm under a integer constraint setting.

\section{Differentiable Random Integers Using STGS}
Although Straight-Through Gumbel-Softmax (STGS) provides a good reparameterization for discrete random variables, directly using discrete variables from STGS to represent integer variables will result in high dimensionality. Thereby, in this section, we will discuss the structure of integer variables and propose a simple linear map trick to reparameterize an integer variable from STGS.

Integer variables usually appear in the industrial setting for the modeling of integer-only or binary decision variables. Integer-only variables can be the number of products or the number of a device's modes. Binary decision variables can be whether to open/close a connection or occupy/release a resource. Despite the discreteness, integer and continuous variables share one thing in common: comparability, or more mathematically, form a totally ordered set. This is the key that makes integer variables distinct from discrete counterparts and allows us to treat integer variables like continuous instead of discrete/categorical variables.

Precisely, a totally ordered set $S$ is a set that satisfies reflexivity ($a\leq a,~\forall~a\in S$), antisymmetry ($a\leq b~\&~b\leq a\Rightarrow a=b$), transitivity ($a\leq b~\&~b\leq c\Rightarrow a\leq c$), and comparability ($\forall~a,b~\in~S,~\text{either}~a\leq b~\text{or}~b\leq a$). It is obvious that a continuous variable (e.g.~$x\in~[-1,1]$) and a integer variable (e.g.~$x\in\{0,1,2\}$) range in totally order sets. On the other hand, a discrete/categorical set does not admit a natural comparison; e.g.,~for~$\{\text{dog},\text{cat}\}$, cat and dog are incomparable. Thus, the comparability fails, and neither do the other properties of the totally ordered set.

Due to the lack of comparability, we usually use an one-hot encoding to represent a discrete/categorical variable; e.g., for $\{\text{dog},\text{cat}\}$, we have dog$\rightarrow[1,0]$ and cat$\rightarrow[0,1]$. The encoding dimension inevitably depends on the size of the range, and this becomes problematic when the range is of large size or when there are multiple discrete variables. On the other hand, with the comparability, an integer variable's effective dimension is only one, regardless of the size of its range; e.g., both variables with range $\{0,1\}$ or $\{0,...,32\}$ need only one dimension to represent. Since the integer variable does not require a one-hot encoding, once being generated, we can treat it like a continuous variable in the subsequent computation. This greatly reduces the computational overhead and is the key to designing a SAC algorithm under integer actions.

The proposed integer reparameterization is as follows. Recall that the Straight-Through Gumbel Softmax, Eq.~\eqref{eq:stgs}, generates a differentiable one-hot random vector $D_{\text{STGS}}$. Then, $\underset{i}{\arg\max} D_{\text{STGS}}[i]$, the index/argument of $1$ in $D_{\text{STGS}}$, is an integer random variable following the distribution formed by $\sm(\text{logit}_\theta)$, as the LHS of Eq.~\eqref{eq:gumbel-max}. Thereby, once establishing an differentiable mapping from $D_{\text{STGS}}$ to $\underset{i}{\arg\max} D_{\text{STGS}}[i]$, we can have an differentiable random integer parameterized by $\theta$. Such a differentiable mapping can be constructed by a simple linear map:
\begin{equation}
    \text{Ind}_\theta=\langle [0,1,...,n-1],D_{\text{STGS}} \rangle,
\label{eq:integer-para}
\end{equation}
where $\langle\cdot,\cdot\rangle$ is an inner product and $D_{\text{STGS}}$ is assumed to be $n$-dimensional. 

Eq.~\eqref{eq:integer-para} gives a parameterized random integer $\text{Ind}_\theta$ with value equaling to $\underset{i\in[0,...,n-1]}{\arg\max} D_{\text{STGS}}[i]$. We apply this to the SAC under integer actions as follows. Since each integer action is of effective dimension 1 (in a euclidean space), the action dimension is precisely the number of integer action variables. Thereby, the input/output dimension of the actor network (policy) and the critic network (Q function) is the same as their continuous action counterparts. The only difference lies in the policy network design, where we restrict the output to have an integer structure using Eq.~\eqref{eq:integer-para}
while the continuous action version doesn't have such a restriction. 

\section{Related Work}
Reinforcement learning (RL) with discrete action space has been studied in different aspects over the literature. Due to the high dimensionality of a discretized action space, it has been of great interest to scale RL to large spaces using a neural network. Examples of this line of research include sequentially predicting an action dimension at a time \citep{metz2017discrete}, training a Q function for each action dimension \citep{tavakoli2018action}, and factorizing the action across dimension using the natural ordering
between discrete actions \citep{tang2020discretizing} or normalizing flows \citep{tang2018boosting,delalleau2019discrete}. Various techniques have been proposed in different settings. When the action space admits a continuity structure, a discrete action can be derived by rounding a continuous one \citet{Hasselt2009discrete}. When the action is represented by a one-hot vector, discrete soft actor-critic \citep{christodoulou2019sac-disc} and Gumbel-Softmax \citep{yan2021discrete} have been taken as key techniques to tackle RL problems.

Our work and the literature share the same motivation as to apply RL to large action spaces. Although our work includes techniques that have been studied (e.g., SAC and Gumbel-Softmax), we reformulate them using a more efficient representation. For example, the one-hot encoding is required in discrete soft actor-critic \citep{christodoulou2019sac-disc} and Gumbel-Softmax for image-text matching \citep{yan2021discrete}. However, as proposed in Eq.~\eqref{eq:integer-para}, our integer representation is much simpler than a one-hot encoding. The key reason for having such a reduction is that an integer set admits a continuity structure. Although this continuity observation has been used to justify the rounding technique \citep{Hasselt2009discrete}, rounding gives only a non-differentiable integer, rather than a differentiable one as required in this work. Hence, our work is different from the literature in the sense that we re-organize the techniques and give a simpler representation of differentiable random integers.

\section{Experiments and Discussions}
We evaluate our integer reparameterization for SAC on two  OpenAI Gym environments \citep{brockman2016openai}: PyBullet robot control \citep{coumans2021bullet} and PowerGym power network control\cite{fan2021powergym}. The results show the proposed SAC under integer actions achieve similar performance as the continuous action version in Pybullet and outperform PPO \citep{schulman2017ppo} algorithm in PowerGym. Hence this shows the effectiveness of SAC under the proposed integer reparameterization.

\subsection{PyBullet Environment}
PyBullet is an open-source version of the MuJoCo environment \citep{todorov2012mujoco} for robot control tasks. Compared to MuJoCo, PyBullet is license-free and is reported to be more difficult to solve \citep{tan2018bullet-exp,fayad2021bullet-exp}. Since the actions in PyBullet are continuous and bounded, we can derive the integer action version of PyBullet by discretizing the actions. Hence, we adopt the PyBullet environment to evaluate the performance discrepancy between integer actions and continuous actions for robot control tasks.

Suppose the dimension of a PyBullet continuous action space is $K$. Because each dimension of a PyBullet action is bounded in an interval $I=[-1,1]$, we can discretize each dimension as follows. Suppose the interval $I$ is discretized into $N\geq 2$ integers as $I_\text{d}=\{0,~1,...,N-1\}$. After having an integer action $a\in I_\text{d}$, $a$ is embedded into $I$ by 
$$\mathcal{T}(a)=a\frac{2}{N-1}-1\in I.$$
Such discretization produces a discrete action space with size $N^K$ where $K$ is the euclidean dimension of a PyBullet action space. The size quickly becomes intractable if using a one-hot encoding for the discrete actions. Even if using a multi-hot encoding, the size of a multi-hot vector, $NK$, is still undesirable for the implementation of the actor and critic networks. On the other hand, if we apply the proposed integer reparameterization, Eq.~\eqref{eq:integer-para}, for each dimension of a PyBullet action, then we recover a $K$-dimensional integer vector because each integer action is of effective dimension 1. Hence, the discretized environment is highly nontrivial under one-hot and multi-hot encodings but is feasible under the proposed integer reparameterization.

\begin{figure}[!ht]
    \centering
    \includegraphics[width=0.4\textwidth]{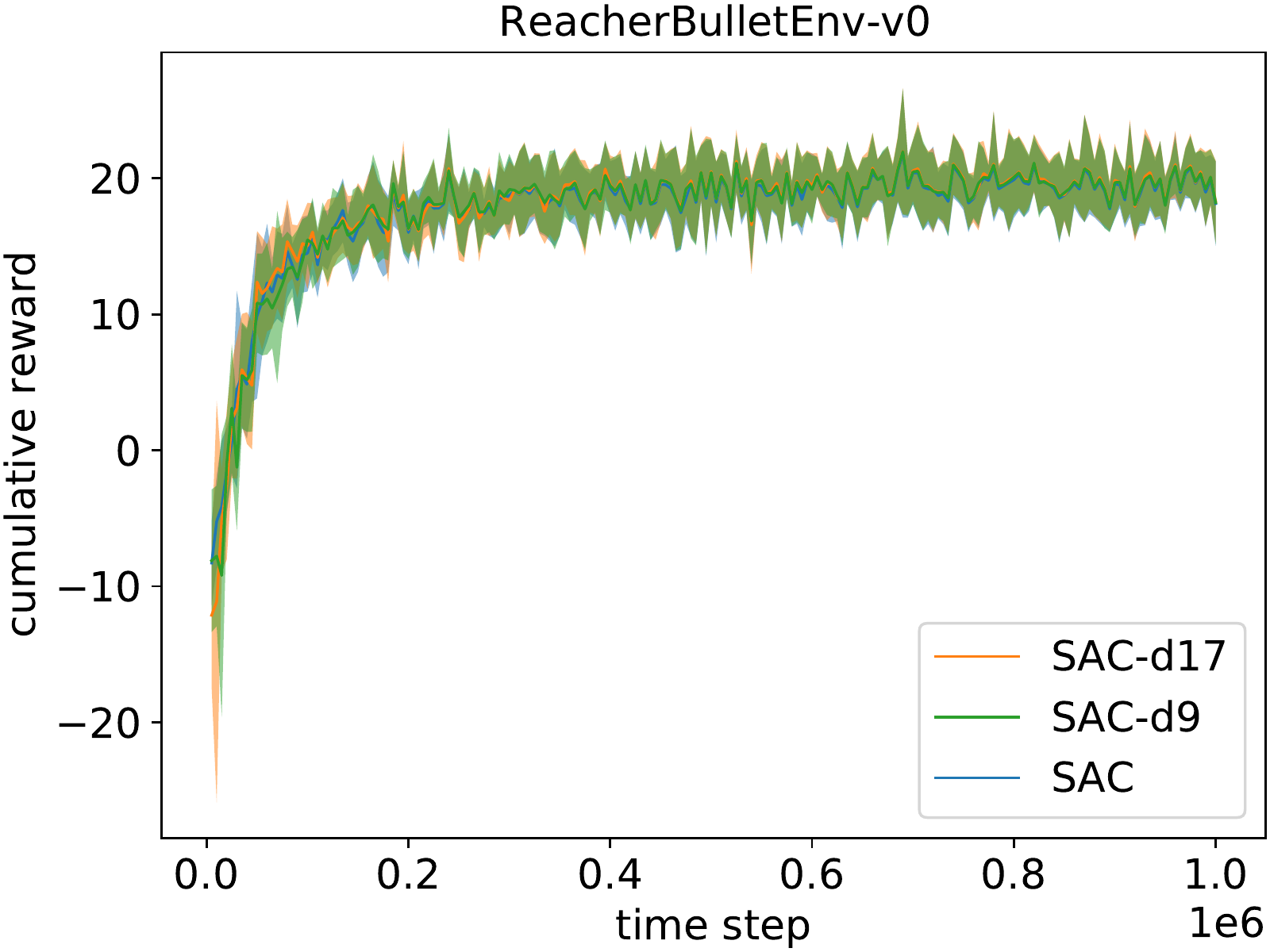}
    \includegraphics[width=0.4\textwidth]{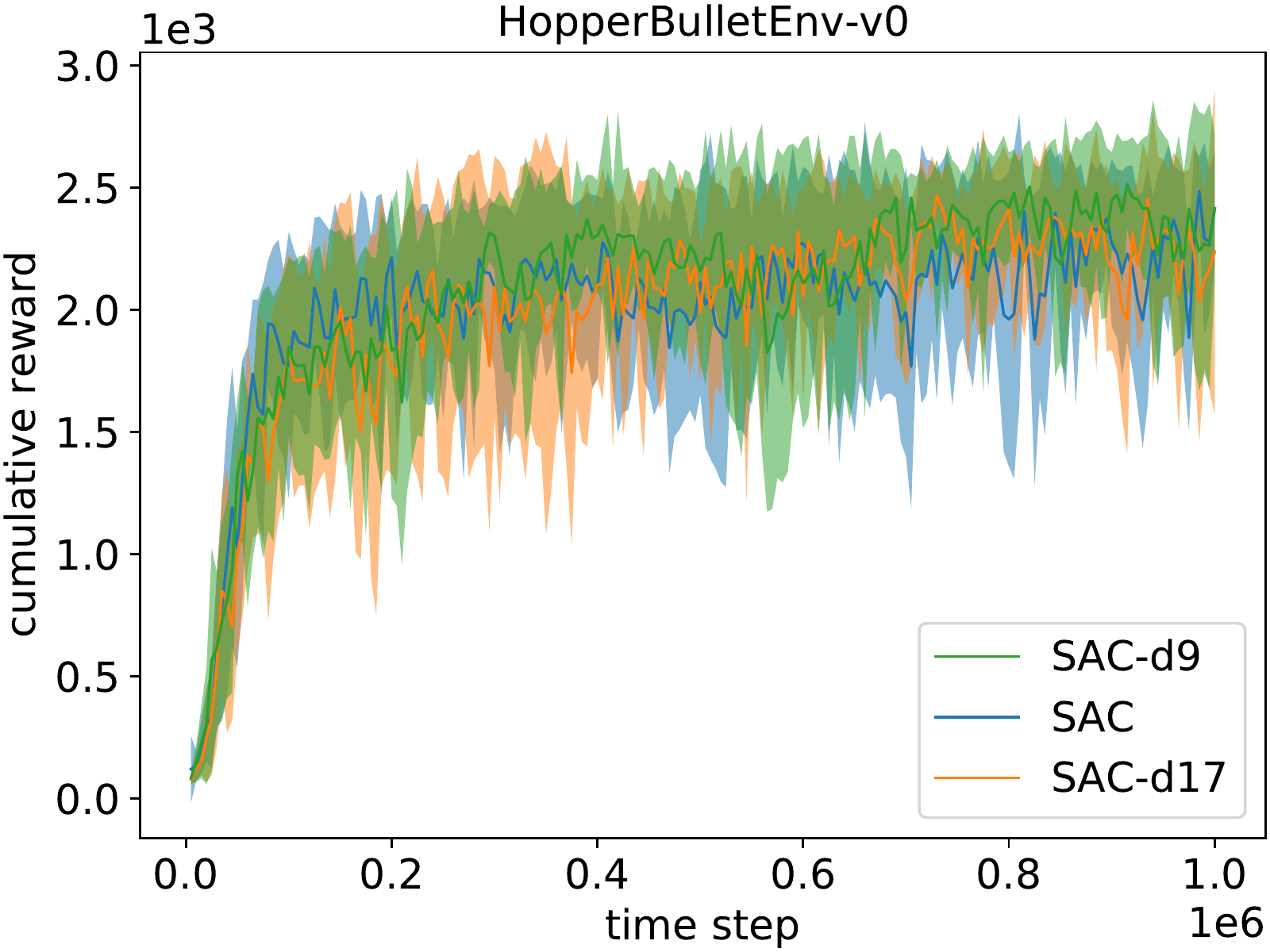}
    \includegraphics[width=0.4\textwidth]{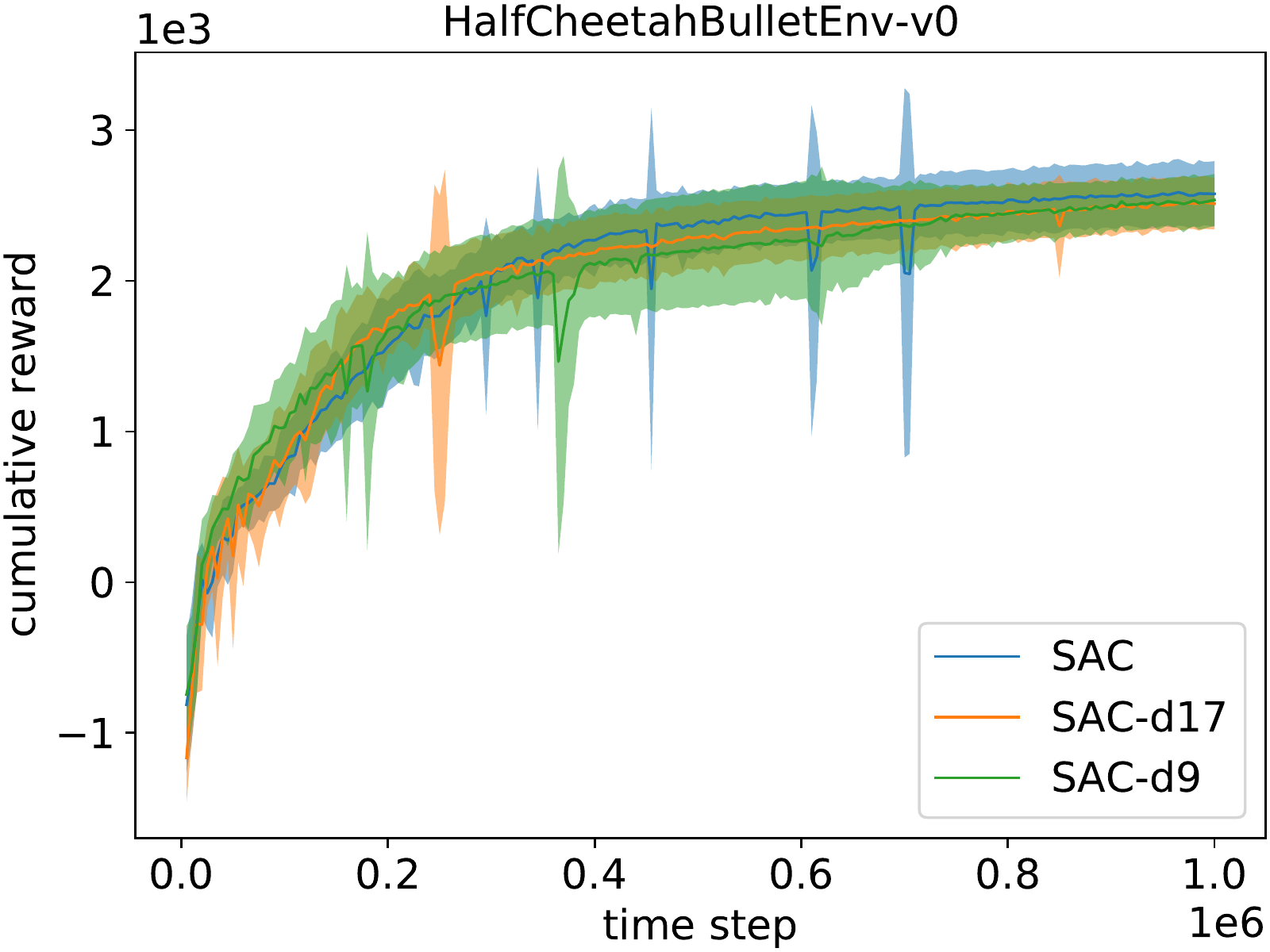}
    \includegraphics[width=0.4\textwidth]{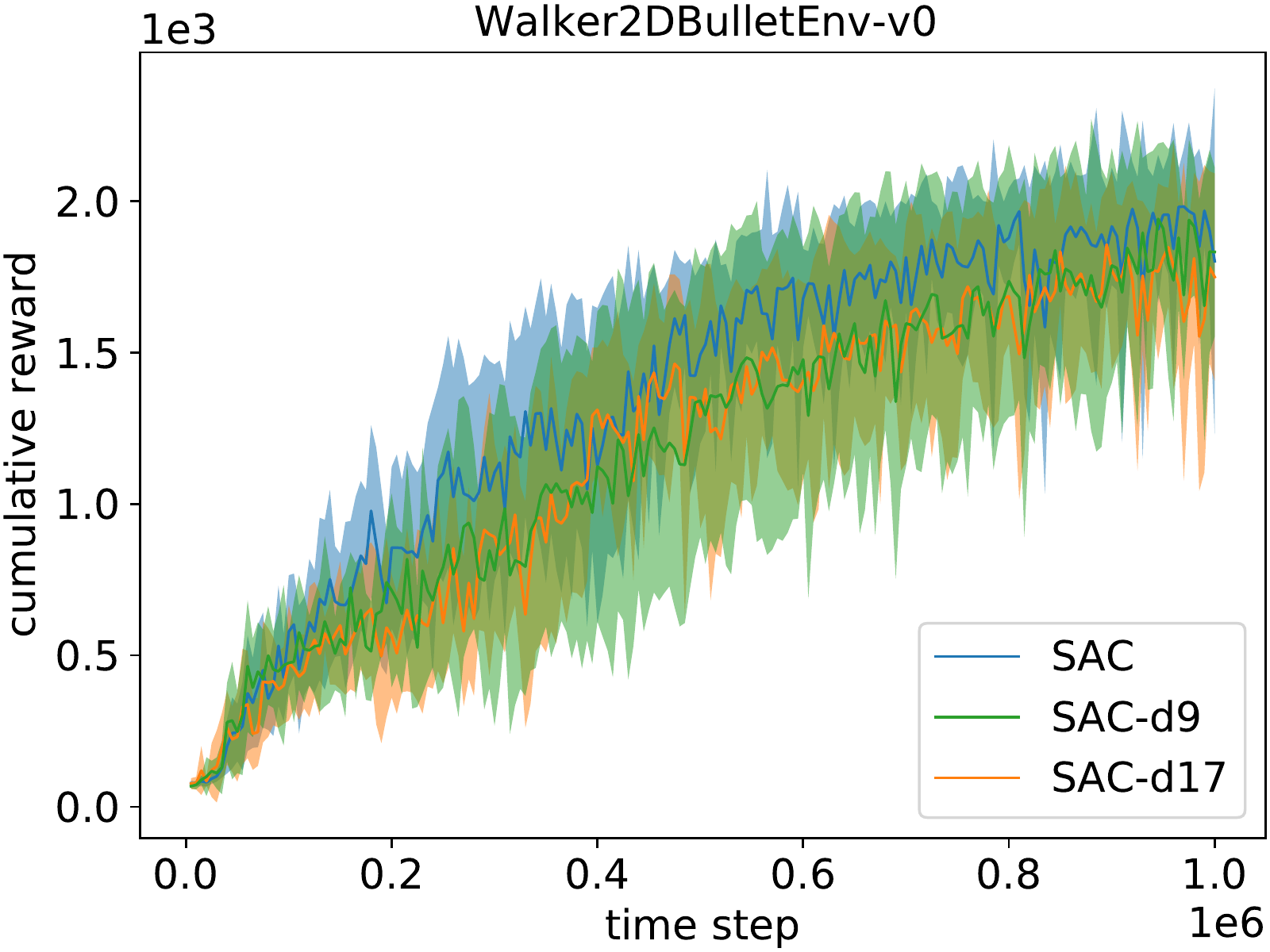}
    \caption{Experiments on PyBullet. The solid lines are the averages and the shaded regions are the standard deviation. d9 and d17 mean to discretize each dimension of the action space into 9 and 17 integers, respectively.}
    \label{fig:bullet-exp}
\end{figure}

\begin{table}[!ht]
    \centering
    \begin{tabular}{lllll}
        \hline
        \hline
        \textbf{Environment}  & Reacher & Hopper & HalfCheetah & Walker2D \\ \hline
        \textbf{Temperature} & 0.005 & 0.05 & 0.02 & 0.06\\ 
        \hline
        \hline
    \end{tabular}
    \caption{Temperature parameters for PyBullet environments.}
    \label{tbl:bullet-temp}
\end{table}
We run SAC in continuous or discretized (integer) actions in four PyBullet environments: Reacher, Hopper, HalfCheetah, Walker2D. Both (cont. or int.) use the same temperature parameters from Table~\ref{tbl:bullet-temp}. Figure~\ref{fig:bullet-exp} shows SAC with continuous or integer actions share very similar performances. Note SAC with continuous actions is known to perform well on PyBullet tasks \citep{raffin2021bullet-exp, fayad2021bullet-exp} in the sense that it outperforms PPO. Therefore, this shows our SAC with integer actions is among the best and certifies the usefulness of the proposed integer reparameterization trick. 

\subsection{PowerGym Environment}
We demonstrate a use case of the proposed integer constraint SAC in an industrial setting. PowerGym \citep{fan2021powergym} is an open-source reinforcement learning environment for Volt-Var control in power distribution systems. By controlling the circuit elements such as capacitors, regulators, and batteries, the Volt-Var control aims to minimize the voltage violations, control loss, and power loss subject to physical networked constraints and device constraints.

The device constraints restrict the decisions on the controllers to be integers. As shown in Table~\ref{tbl:act_space}, a capacitor's status is binary (on/off), a regulator's tap number is an integer in finite range, and a battery's discharge power is also represented by an integer in finite range. Hence, PowerGym is a good example of a practical environment with integer actions.

\begin{table}[!ht]
    \centering
    \begin{tabular}{l|ll}
        \hline
        \hline
        \textbf{Variable} & \textbf{Type} & \textbf{Range} \\ \hline
        Capacitor status & disc. & $\{0,1\}$ \\
        Regulator tap number & disc. & $\{0,...,N_{\text{reg\_act}}-1\}$ \\
        Discharge power & disc. & $\{0,...,N_{\text{bat\_act}}-1\}$ \\
        \hline
        \hline
    \end{tabular}
    \caption{Action Space. $N_{\text{reg\_act}}$ and $N_{\text{bat\_act}}$ are 33 by default.}
    \label{tbl:act_space}
\end{table}

\begin{table}[!ht]
    \centering
    \begin{tabular}{l|lll}
        \hline
        \hline
        System  & \# Caps & \# Regs & \# Bats\\ \hline
        13Bus  & 2 & 3 & 1\\
        34Bus  & 2 & 6 & 2\\
        123Bus  & 4 & 7 & 4\\
        8500Node  & 10 & 12 & 10\\ \hline
        \hline
    \end{tabular}
    \caption{Specifications of PowerGym environment.}
    \label{tbl:sys}
\end{table}

\begin{figure}[!ht]
    \centering
    \includegraphics[width=0.4\textwidth]{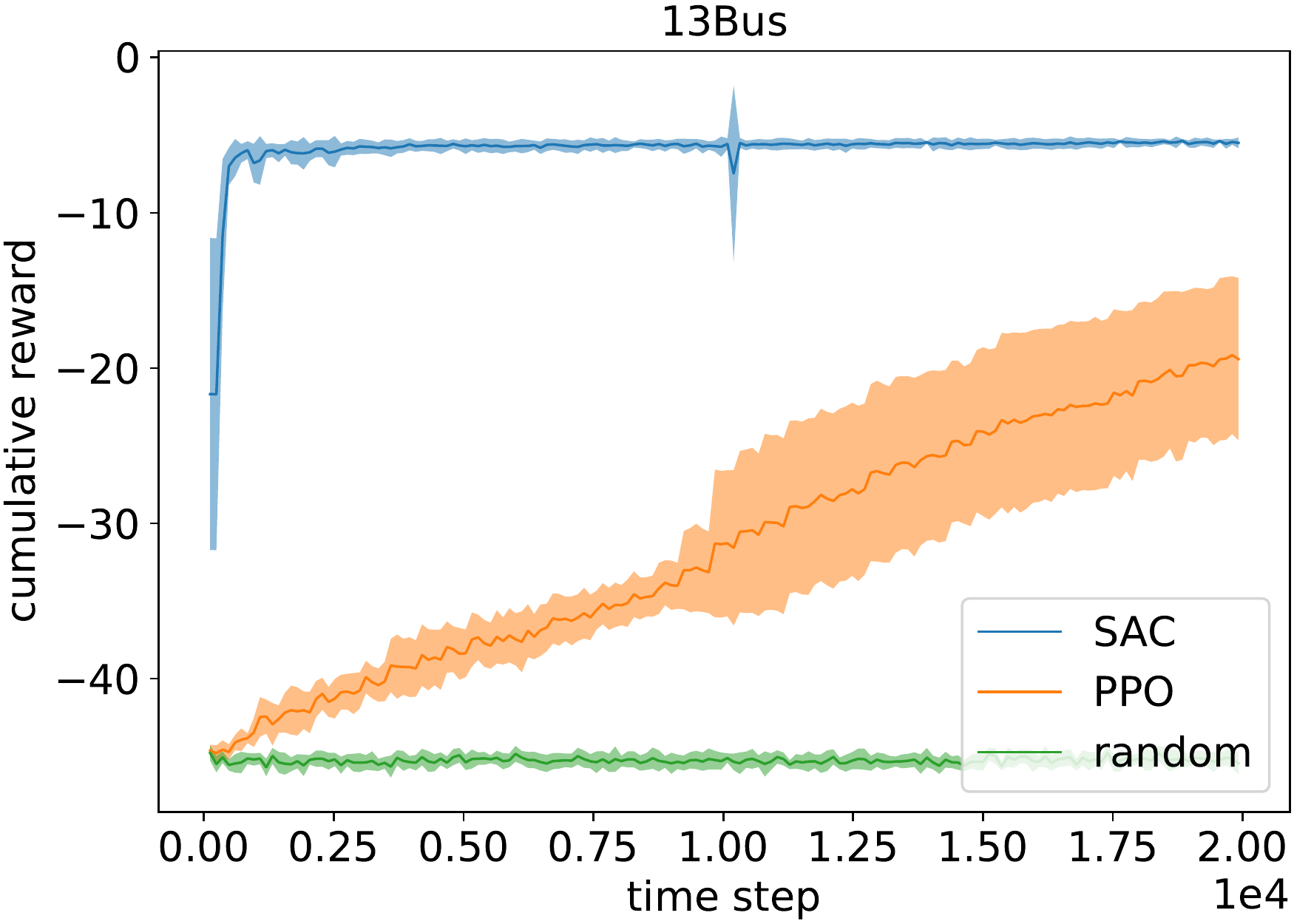}
    \includegraphics[width=0.4\textwidth]{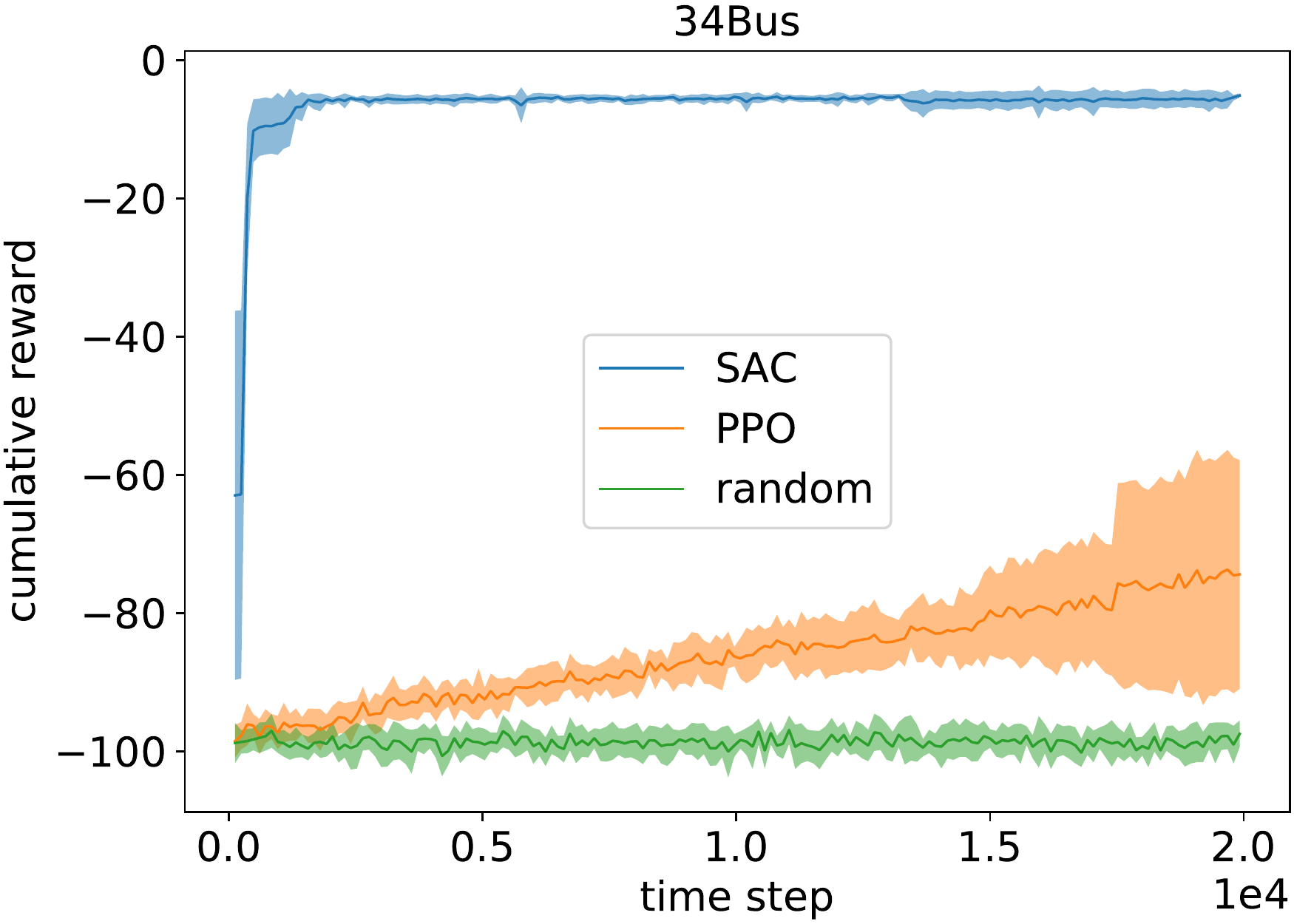}
    \includegraphics[width=0.4\textwidth]{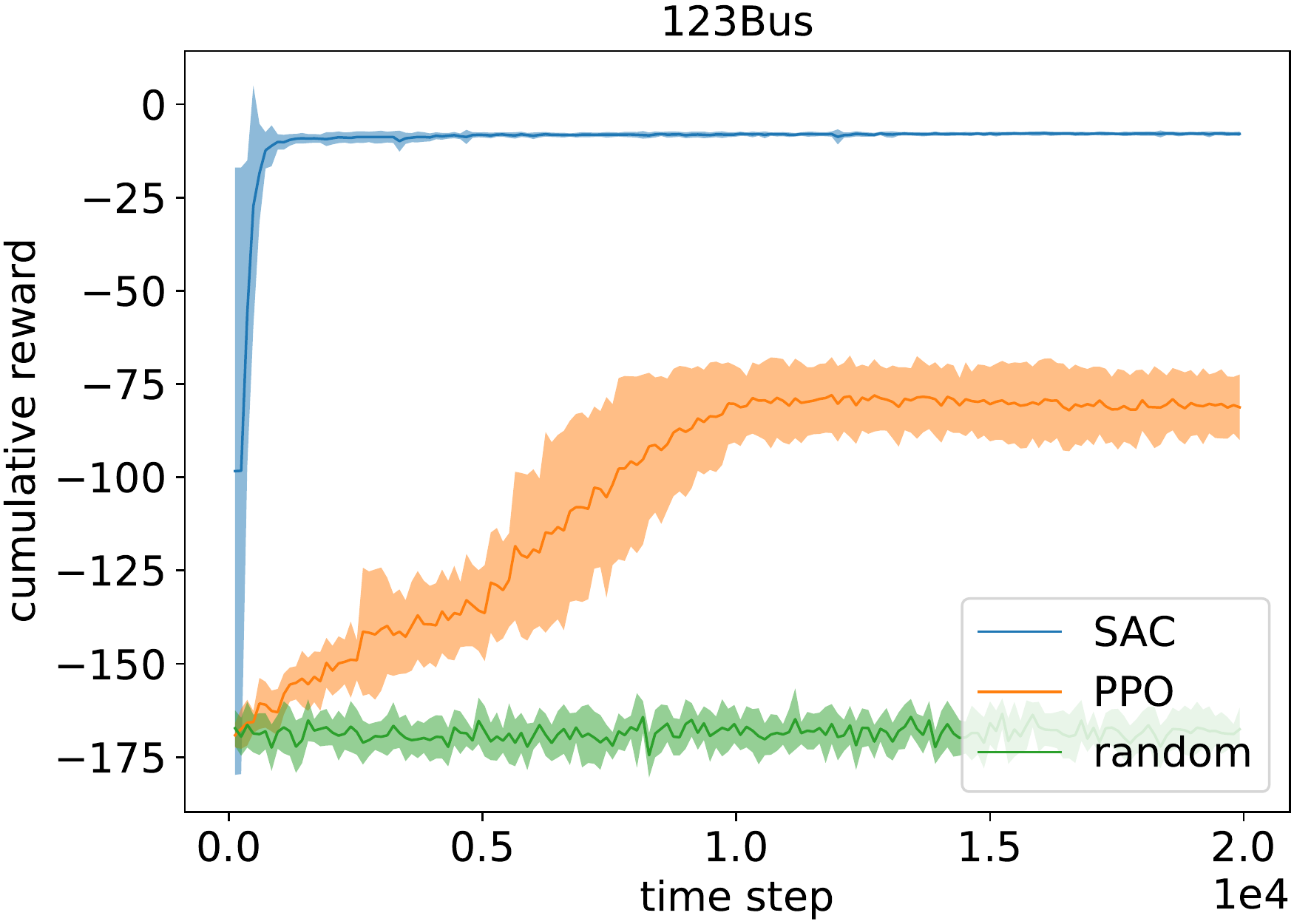}
    \includegraphics[width=0.4\textwidth]{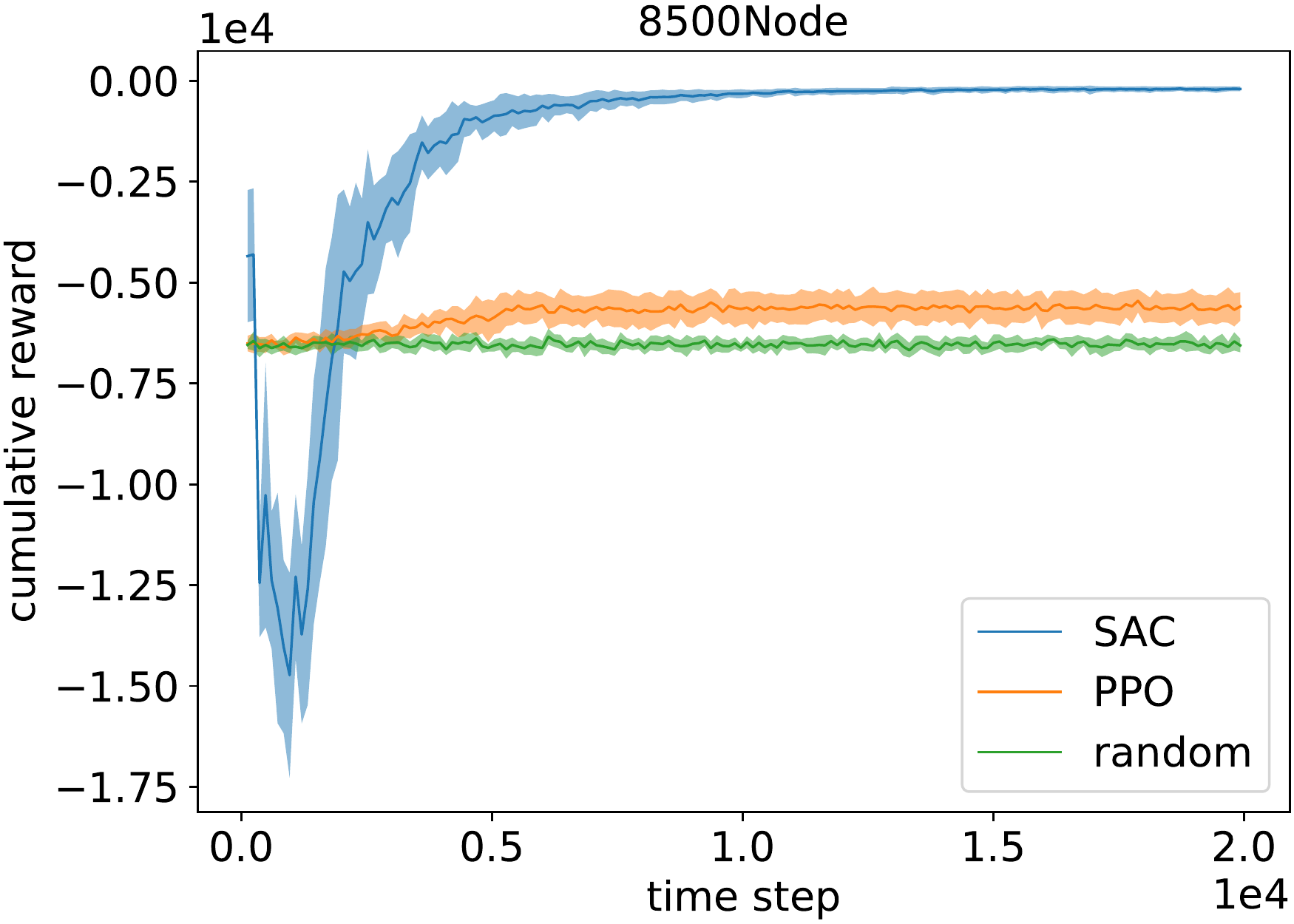}
    \caption{Experiments on PowerGym. "random" denotes a uniform random policy that takes actions uniformly at random from the action space.}
    \label{fig:powergym-exp}
\end{figure}

We evaluate SAC with integer reparameterization on four PowerGym environments, as shown in Table~\ref{tbl:sys}. Since the actions are pure integers, we cannot have a comparison between continuous and discretized actions as Figure~\ref{fig:bullet-exp}. Instead, take PPO \citep{schulman2017ppo} as a benchmark against our SAC under integer actions. Note that the literature always implements SAC with reparameterization and PPO without reparameterization \citep{haarnoja2018sac,raffin2021bullet-exp, fayad2021bullet-exp}. This is because PPO is based on the REINFORCE estimator \citep{glynn1990reinforce} and does not need differentiable random actions. Following this convention, in this experiment, SAC is equipped with integer reparameterization while PPO is using REINFORCE.

Figure~\ref{fig:powergym-exp} shows that SAC outperforms PPO in all four environments. This demonstrates the usefulness of the proposed integer reparameterization for SAC. In addition, because PPO does a clipping at its policy gradient while SAC doesn't, PPO starts near the random policy and improves steadily but slowly. On the other hand, SAC is susceptible to the accuracy of the Q function, so it deviates from the random policy in the early steps but improves quickly as the Q function becomes more accurate, which explains the swing of cumulative reward in the 8500Node experiment.

\section{Conclusion}
Motivated by industrial scenarios, we study reinforcement learning under integer actions. We approach the problem by incorporating the Soft Actor-Critic (SAC) algorithm with an integer reparameterization on actions. Although the Gumbel-Softmax is capable of modeling discrete random variables, the discreteness of the integer actions can be simplified using the comparability property. The proposed integer reparameterization is based on Gumbel-Softmax followed by a linear mapping for output dimension reduction. Experiments show that the proposed SAC under integer actions is as good as the continuous action version on robot control tasks and can outperform Proximal Policy Optimization on power distribution tasks. Hence, it suggests the usefulness of the proposed integer reparameterization on the SAC algorithm.
\bibliographystyle{unsrtnat}
\bibliography{mybib}

\begin{thebibliography}{38}
\providecommand{\natexlab}[1]{#1}
\providecommand{\url}[1]{\texttt{#1}}
\expandafter\ifx\csname urlstyle\endcsname\relax
  \providecommand{\doi}[1]{doi: #1}\else
  \providecommand{\doi}{doi: \begingroup \urlstyle{rm}\Url}\fi

\bibitem[Mnih et~al.(2013)Mnih, Kavukcuoglu, Silver, Graves, Antonoglou,
  Wierstra, and Riedmiller]{mnih2013atari}
Volodymyr Mnih, Koray Kavukcuoglu, David Silver, Alex Graves, Ioannis
  Antonoglou, Daan Wierstra, and Martin Riedmiller.
\newblock Playing atari with deep reinforcement learning.
\newblock \emph{arXiv preprint arXiv:1312.5602}, 2013.

\bibitem[Todorov et~al.(2012)Todorov, Erez, and Tassa]{todorov2012mujoco}
Emanuel Todorov, Tom Erez, and Yuval Tassa.
\newblock Mujoco: A physics engine for model-based control.
\newblock In \emph{2012 IEEE/RSJ International Conference on Intelligent Robots
  and Systems}, pages 5026--5033. IEEE, 2012.

\bibitem[Levine et~al.(2016)Levine, Finn, Darrell, and
  Abbeel]{levine2016robotics}
Sergey Levine, Chelsea Finn, Trevor Darrell, and Pieter Abbeel.
\newblock End-to-end training of deep visuomotor policies.
\newblock \emph{The Journal of Machine Learning Research}, 17\penalty0
  (1):\penalty0 1334--1373, 2016.

\bibitem[Leurent(2018)]{highway-env}
Edouard Leurent.
\newblock An environment for autonomous driving decision-making.
\newblock \url{https://github.com/eleurent/highway-env}, 2018.

\bibitem[Marot et~al.(2021)Marot, Donnot, Dulac-Arnold, Kelly, O'Sullivan,
  Viebahn, Awad, Guyon, Panciatici, and Romero]{marot2021power-env}
Antoine Marot, Benjamin Donnot, Gabriel Dulac-Arnold, Adrian Kelly, A{\"\i}dan
  O'Sullivan, Jan Viebahn, Mariette Awad, Isabelle Guyon, Patrick Panciatici,
  and Camilo Romero.
\newblock Learning to run a power network challenge: a retrospective analysis.
\newblock \emph{arXiv preprint arXiv:2103.03104}, 2021.

\bibitem[Zoltners and Sinha(1980)]{zoltners1980integer}
Andris~A Zoltners and Prabhakant Sinha.
\newblock Integer programming models for sales resource allocation.
\newblock \emph{Management Science}, 26\penalty0 (3):\penalty0 242--260, 1980.

\bibitem[Wang and Usher(2005)]{wang2005scheduling}
Yi-Chi Wang and John~M Usher.
\newblock Application of reinforcement learning for agent-based production
  scheduling.
\newblock \emph{Engineering Applications of Artificial Intelligence},
  18\penalty0 (1):\penalty0 73--82, 2005.

\bibitem[Silver et~al.(2014)Silver, Lever, Heess, Degris, Wierstra, and
  Riedmiller]{silver2014ddpg}
David Silver, Guy Lever, Nicolas Heess, Thomas Degris, Daan Wierstra, and
  Martin Riedmiller.
\newblock Deterministic policy gradient algorithms.
\newblock In \emph{International conference on machine learning}, pages
  387--395. PMLR, 2014.

\bibitem[Haarnoja et~al.(2018)Haarnoja, Zhou, Abbeel, and
  Levine]{haarnoja2018sac}
Tuomas Haarnoja, Aurick Zhou, Pieter Abbeel, and Sergey Levine.
\newblock Soft actor-critic: Off-policy maximum entropy deep reinforcement
  learning with a stochastic actor.
\newblock In \emph{International conference on machine learning}, pages
  1861--1870. PMLR, 2018.

\bibitem[Fujimoto et~al.(2018)Fujimoto, Hoof, and Meger]{fujimoto2018td3}
Scott Fujimoto, Herke Hoof, and David Meger.
\newblock Addressing function approximation error in actor-critic methods.
\newblock In \emph{International Conference on Machine Learning}, pages
  1587--1596. PMLR, 2018.

\bibitem[Glynn(1990)]{glynn1990reinforce}
Peter~W Glynn.
\newblock Likelihood ratio gradient estimation for stochastic systems.
\newblock \emph{Communications of the ACM}, 33\penalty0 (10):\penalty0 75--84,
  1990.

\bibitem[Schulman et~al.(2015)Schulman, Levine, Abbeel, Jordan, and
  Moritz]{schulman2015trpo}
John Schulman, Sergey Levine, Pieter Abbeel, Michael Jordan, and Philipp
  Moritz.
\newblock Trust region policy optimization.
\newblock In \emph{International conference on machine learning}, pages
  1889--1897. PMLR, 2015.

\bibitem[Schulman et~al.(2017)Schulman, Wolski, Dhariwal, Radford, and
  Klimov]{schulman2017ppo}
John Schulman, Filip Wolski, Prafulla Dhariwal, Alec Radford, and Oleg Klimov.
\newblock Proximal policy optimization algorithms.
\newblock \emph{arXiv preprint arXiv:1707.06347}, 2017.

\bibitem[Schulman et~al.(2016)Schulman, Moritz, Levine, Jordan, and
  Abbeel]{Schulman2015advantage}
John Schulman, Philipp Moritz, Sergey Levine, Michael~I. Jordan, and Pieter
  Abbeel.
\newblock High-dimensional continuous control using generalized advantage
  estimation.
\newblock In Yoshua Bengio and Yann LeCun, editors, \emph{4th International
  Conference on Learning Representations, {ICLR} 2016, San Juan, Puerto Rico,
  May 2-4, 2016, Conference Track Proceedings}, 2016.

\bibitem[Wu et~al.(2018)Wu, Rajeswaran, Duan, Kumar, Bayen, Kakade, Mordatch,
  and Abbeel]{wu2018variance}
Cathy Wu, Aravind Rajeswaran, Yan Duan, Vikash Kumar, Alexandre~M Bayen, Sham
  Kakade, Igor Mordatch, and Pieter Abbeel.
\newblock Variance reduction for policy gradient with action-dependent
  factorized baselines.
\newblock In \emph{International Conference on Learning Representations}, 2018.

\bibitem[Jang et~al.(2017)Jang, Gu, and Poole]{Jang2017gumbel}
Eric Jang, Shixiang Gu, and Ben Poole.
\newblock Categorical reparameterization with gumbel-softmax.
\newblock In \emph{5th International Conference on Learning Representations,
  {ICLR} 2017, Toulon, France, April 24-26, 2017, Conference Track
  Proceedings}. OpenReview.net, 2017.

\bibitem[Maddison et~al.(2017)Maddison, Mnih, and Teh]{Chris2017gumbel}
Chris~J. Maddison, Andriy Mnih, and Yee~Whye Teh.
\newblock The concrete distribution: {A} continuous relaxation of discrete
  random variables.
\newblock In \emph{5th International Conference on Learning Representations,
  {ICLR} 2017, Toulon, France, April 24-26, 2017, Conference Track
  Proceedings}. OpenReview.net, 2017.

\bibitem[Bengio et~al.(2013)Bengio, L{\'e}onard, and Courville]{bengio2013st}
Yoshua Bengio, Nicholas L{\'e}onard, and Aaron Courville.
\newblock Estimating or propagating gradients through stochastic neurons for
  conditional computation.
\newblock \emph{arXiv preprint arXiv:1308.3432}, 2013.

\bibitem[Chung et~al.(2017)Chung, Ahn, and Bengio]{Chung2017st}
Junyoung Chung, Sungjin Ahn, and Yoshua Bengio.
\newblock Hierarchical multiscale recurrent neural networks.
\newblock In \emph{5th International Conference on Learning Representations,
  {ICLR} 2017, Toulon, France, April 24-26, 2017, Conference Track
  Proceedings}. OpenReview.net, 2017.

\bibitem[Coumans and Bai(2016--2021)]{coumans2021bullet}
Erwin Coumans and Yunfei Bai.
\newblock Pybullet, a python module for physics simulation for games, robotics
  and machine learning.
\newblock \url{http://pybullet.org}, 2016--2021.

\bibitem[Fan et~al.(2021)Fan, Lee, and Wang]{fan2021powergym}
Ting-Han Fan, Xian~Yeow Lee, and Yubo Wang.
\newblock Powergym: A reinforcement learning environment for volt-var control
  in power distribution systems.
\newblock \emph{arXiv preprint arXiv:2109.03970}, 2021.

\bibitem[Sutton et~al.(2000)Sutton, McAllester, Singh, and
  Mansour]{sutton2000policy}
Richard~S Sutton, David~A McAllester, Satinder~P Singh, and Yishay Mansour.
\newblock Policy gradient methods for reinforcement learning with function
  approximation.
\newblock In \emph{Advances in neural information processing systems}, pages
  1057--1063, 2000.

\bibitem[Williams(1992)]{williams1992reinforce}
Ronald~J Williams.
\newblock Simple statistical gradient-following algorithms for connectionist
  reinforcement learning.
\newblock \emph{Machine learning}, 8\penalty0 (3):\penalty0 229--256, 1992.

\bibitem[Paulus et~al.(2021)Paulus, Maddison, and
  Krause]{paulus2021raoblackwellizing}
Max~B Paulus, Chris~J. Maddison, and Andreas Krause.
\newblock Rao-blackwellizing the straight-through gumbel-softmax gradient
  estimator.
\newblock In \emph{International Conference on Learning Representations}, 2021.

\bibitem[Maddison et~al.(2014)Maddison, Tarlow, and Minka]{chris2014gumbel}
Chris~J Maddison, Daniel Tarlow, and Tom Minka.
\newblock A* sampling.
\newblock In Z.~Ghahramani, M.~Welling, C.~Cortes, N.~Lawrence, and K.~Q.
  Weinberger, editors, \emph{Advances in Neural Information Processing
  Systems}, volume~27. Curran Associates, Inc., 2014.

\bibitem[Hinton et~al.(2012)Hinton, Srivastava, and Swersky]{hinton2012neural}
Geoffrey Hinton, Nitsh Srivastava, and Kevin Swersky.
\newblock Neural networks for machine learning.
\newblock \emph{Coursera, video lectures}, 2012.

\bibitem[Metz et~al.(2017)Metz, Ibarz, Jaitly, and Davidson]{metz2017discrete}
Luke Metz, Julian Ibarz, Navdeep Jaitly, and James Davidson.
\newblock Discrete sequential prediction of continuous actions for deep rl.
\newblock \emph{arXiv preprint arXiv:1705.05035}, 2017.

\bibitem[Tavakoli et~al.(2018)Tavakoli, Pardo, and
  Kormushev]{tavakoli2018action}
Arash Tavakoli, Fabio Pardo, and Petar Kormushev.
\newblock Action branching architectures for deep reinforcement learning.
\newblock In \emph{Proceedings of the AAAI Conference on Artificial
  Intelligence}, volume~32, 2018.

\bibitem[Tang and Agrawal(2020)]{tang2020discretizing}
Yunhao Tang and Shipra Agrawal.
\newblock Discretizing continuous action space for on-policy optimization.
\newblock In \emph{Proceedings of the AAAI Conference on Artificial
  Intelligence}, volume~34, pages 5981--5988, 2020.

\bibitem[Tang and Agrawal(2018)]{tang2018boosting}
Yunhao Tang and Shipra Agrawal.
\newblock Boosting trust region policy optimization by normalizing flows
  policy.
\newblock \emph{arXiv preprint arXiv:1809.10326}, 2018.

\bibitem[Delalleau et~al.(2019)Delalleau, Peter, Alonso, and
  Logut]{delalleau2019discrete}
Olivier Delalleau, Maxim Peter, Eloi Alonso, and Adrien Logut.
\newblock Discrete and continuous action representation for practical rl in
  video games.
\newblock \emph{arXiv preprint arXiv:1912.11077}, 2019.

\bibitem[van Hasselt and Wiering(2009)]{Hasselt2009discrete}
Hado van Hasselt and Marco~A. Wiering.
\newblock Using continuous action spaces to solve discrete problems.
\newblock In \emph{2009 International Joint Conference on Neural Networks},
  pages 1149--1156, 2009.

\bibitem[Christodoulou(2019)]{christodoulou2019sac-disc}
Petros Christodoulou.
\newblock Soft actor-critic for discrete action settings.
\newblock \emph{arXiv preprint arXiv:1910.07207}, 2019.

\bibitem[Yan et~al.(2021)Yan, Yu, and Xie]{yan2021discrete}
Shiyang Yan, Li~Yu, and Yuan Xie.
\newblock Discrete-continuous action space policy gradient-based attention for
  image-text matching.
\newblock In \emph{Proceedings of the IEEE/CVF Conference on Computer Vision
  and Pattern Recognition}, pages 8096--8105, 2021.

\bibitem[Brockman et~al.(2016)Brockman, Cheung, Pettersson, Schneider,
  Schulman, Tang, and Zaremba]{brockman2016openai}
Greg Brockman, Vicki Cheung, Ludwig Pettersson, Jonas Schneider, John Schulman,
  Jie Tang, and Wojciech Zaremba.
\newblock Openai gym.
\newblock \emph{arXiv preprint arXiv:1606.01540}, 2016.

\bibitem[Tan et~al.(2018)Tan, Zhang, Coumans, Iscen, Bai, Hafner, Bohez, and
  Vanhoucke]{tan2018bullet-exp}
Jie Tan, Tingnan Zhang, Erwin Coumans, Atil Iscen, Yunfei Bai, Danijar Hafner,
  Steven Bohez, and Vincent Vanhoucke.
\newblock Sim-to-real: Learning agile locomotion for quadruped robots.
\newblock In \emph{Robotics: Science and Systems (RSS)}, 2018.

\bibitem[Fayad and Ibrahim(2021)]{fayad2021bullet-exp}
Ammar Fayad and Majd Ibrahim.
\newblock Behavior-guided actor-critic: Improving exploration via learning
  policy behavior representation for deep reinforcement learning.
\newblock \emph{arXiv preprint arXiv:2104.04424}, 2021.

\bibitem[Raffin et~al.(2021)Raffin, Kober, and Stulp]{raffin2021bullet-exp}
Antonin Raffin, Jens Kober, and Freek Stulp.
\newblock Smooth exploration for robotic reinforcement learning.
\newblock \emph{arXiv preprint arXiv:2005.05719}, 2021.

\end{thebibliography}



\end{document}